\pdfoutput=1

\documentclass[11pt]{article}

\usepackage{acl}
\usepackage{times}
\usepackage{latexsym}
\usepackage{textgreek}
\usepackage{mathtools}
\usepackage{amsmath}
\usepackage{txfonts}
\usepackage{xspace}
\usepackage{booktabs}
\usepackage{multirow}
\usepackage{enumitem} 
\usepackage{lipsum}
\usepackage{algorithm}        
\usepackage{pifont}
\usepackage{subcaption}
\usepackage[noend]{algpseudocode}
\newcommand{\macrof}{$\mathrm{m}$-$\mathrm{F_1}$\xspace}
\newcommand{\microf}{$\mathrm{\muup}$-$\mathrm{F_1}$\xspace}

\newcommand{\cls}{\texttt{\small </s>}\xspace}
\newcommand{\pseudolabel}{\texttt{\small <label\_1>}\xspace}

\usepackage{microtype}

\title{An Exploration of Encoder-Decoder Approaches to \\Multi-Label Classification for Legal and Biomedical Text}

\date{2023}
\author{Yova Kementchedjhieva\thanks{\hspace{0.5em}Equal contribution.} \qquad Ilias Chalkidis$^\ast$ \\
Department of Computer Science, University of Copenhagen, Denmark \\
\texttt{\{yova,ilias.chalkidis\}[at]di.ku.dk}}
\begin{document}
\maketitle

\begin{abstract}

Standard methods for multi-label text classification largely rely on encoder-only pre-trained language models, whereas encoder-decoder models have proven more effective in other classification tasks.
In this study, we compare four methods for multi-label classification, two based on an encoder only, and two based on an encoder-decoder. We carry out experiments on four datasets---two in the legal domain and two in the biomedical domain, each with two levels of label granularity--- and always depart from the same pre-trained model, T5. Our results show that encoder-decoder methods outperform encoder-only methods, with a growing advantage on more complex datasets and labeling schemes of finer granularity. 
Using encoder-decoder models in a non-autoregressive fashion, in particular, yields the best performance overall, so we further study this approach through ablations to better understand its strengths.

\end{abstract}

\section{Introduction}

Multi-label classification constitutes the task of predicting multiple labels for an input as opposed to a single (possibly binary) one. The labels are drawn from a set of up to several hundred classes, often with the added challenge of class imbalance. While the order in which labels are predicted is irrelevant, there can be interdependence between subsets of labels. The task is commonly approached with a classification model based on a pre-trained encoder followed by a multi-output classification head.

Encoder-decoder models, like T5 \cite{raffel-etal-2020}, have taken over recent NLP literature with state-of-the-art results on various tasks, such as question-answering (QA), summarization, single-label classification, etc. \citet{raffel-etal-2020} showed that any given NLP task could be reformulated as a \emph{text-to-text} task and solved with conditional generation, i.e., generating a text sequence that represents the desired output, be that 
a span of text in QA, a text summary, a label descriptor, etc. \citet{liu-etal-2022-t5enc} presented an alternative use of encoder-decoder models for classification tasks in particular, wherein T5's decoder is used in a non-autoregressive fashion to obtain output representations, which are then fed to a classification head. 

The application of encoder-decoder methods to multi-label classification is currently limited to one experiment in the work of \citet{liu-etal-2022-t5enc}, who compare a text-to-text approach and their non-autoregressive approach on a single dataset, including an encoder-only baseline built off of a different pre-trained model, BERT \cite{devlin-etal-2019-bert}. They obtain results favorable to the two encoder-decoder methods, but since the focus of their work is not multi-label classification in particular, their evaluation is insufficient to draw hard conclusions about this task, and analysis on the contribution of different model components to performance on the task is missing altogether.

In this work, we carry out an extensive study of encoder-decoder approaches to multi-label classification. To ensure the thorough and fair evaluation of all methods:

\vspace{-1mm}
\begin{enumerate}[label=(\alph*)]
    \item We experiment on four datasets from two different domains (legal and biomedical), each with two levels of label granularity.
    \item We include four methods for multi-label classification, two encoder-only methods and two encoder-decoder methods.
    \item We conduct preliminary development to determine the best configuration for the application of each method, e.g. choice of label descriptors for the text-to-text approach.
    \item We explore how model size affects performance, by fine-tuning small, base, and large T5 models.
    \item We ablate components of the best performing approach, the non-autoregressive encoder-decoder method of \citet{liu-etal-2022-t5enc}, to better understand its strengths. 
\end{enumerate}
We release our code base to assure reproducibility and let others extend our study by experimenting with new methods and more datasets.\footnote{\url{https://github.com/coastalcph/Multi-Label-Classification-T5}}

\section{Related Work}

Class imbalance is a critical issue in multi-label classification, with researchers searching for the best method to handle rare (less represented) labels.

\paragraph{Encoder-only Approaches}
\citet{snell-etal-2017} introduced the idea of a \textit{prototype} label vector, obtained by averaging over all instances of a given class and used to add inductive bias to their Prototypical Network for multi-label classification. In a similar vein, \citet{Mullenbach2018} developed the Label-Wise Attention Network (LWAN) architecture, in which label-wise document representations are obtained by learning to attend to the most informative input words for each label, using trainable label vectors as keys.

\citet{chalkidis-etal-2020-empirical} systematically studied the effects of different language encoders (CNNs, BIGRUs, BERT) and several variants of LWAN with regards to the representation of prototype labels. Experimenting with three datasets (EURLEX, MIMIC-III, and AMAZON), they showed that better language encoders counter-play the positive effect of the LWAN module, i.e., a standard BI-GRU classifier outperforms CNN-based LWANs \cite{Mullenbach2018}, and a standard BERT outperforms BIGRU-LWAN, respectively.  Moreover, BERT-based LWANs offer minor overall improvements compared to a vanilla BERT classifier, wherein BERT's \textit{CLS} token representation is passed to a classification head \cite{devlin-etal-2019-bert}.

\citet{chalkidis2021-multieurlex} were the first to explore the use of a T5 model for multi-label classification, although they only considered an encoder-only classifier, disregarding the model's decoder. They followed the now standard approach of a classification head on top of the \cls token representation. In experiments with mT5 \cite{xue-etal-2021-mt5}, they showcased improved results compared to XLM-R \cite{conneau-etal-2020-unsupervised} on a newly introduced multilingual dataset, MultiEURLEX. 

\paragraph{Encoder-Decoder Approaches} Text-to-text approaches, which utilize the full encoder-decoder model, have proven effective for binary and single-label classification tasks \cite{raffel-etal-2020, won-etal-2022-flan5}. The key to such approaches are label verbalizers, words in natural language which verbalize the underlying semantics of a given class. Label verbalizers are represented in the embedding space of pre-trained models and in this way benefit from the model pre-training. This can be more optimal especially for few- and zero-shot labels, in comparison to head-based classification methods where randomly initialized parameters have to be learned from scratch. 

\citet{liu-etal-2022-t5enc} presented an alternative use of the full T5 model for non-autoregressive tasks, e.g. single-label and multi-label classification, wherein the decoder is used to obtain label-wise representations informed by the input document, which in turn are fed to label-specific binary classification heads. \citet{liu-etal-2022-t5enc} performed one set of experiments on the EURLEX-57K dataset \cite{chalkidis-etal-2019-large}, in which they compared their non-autoregressive approach to a T5-based text-to-text approach and a standard BERT-based classifier. They found that both T5-based approaches outperformed the encoder-only classifier, the non-autoregressive method performing best. Nonetheless, the encoder-only classifier had less than half the parameters of the T5 model (110M vs 222M). Encoder-decoder approaches thus seem to carry potential for multi-label classification, still with insufficient empirical evidence, however.

\begin{figure*}
    \centering
    \resizebox{\textwidth}{!}{
    \includegraphics{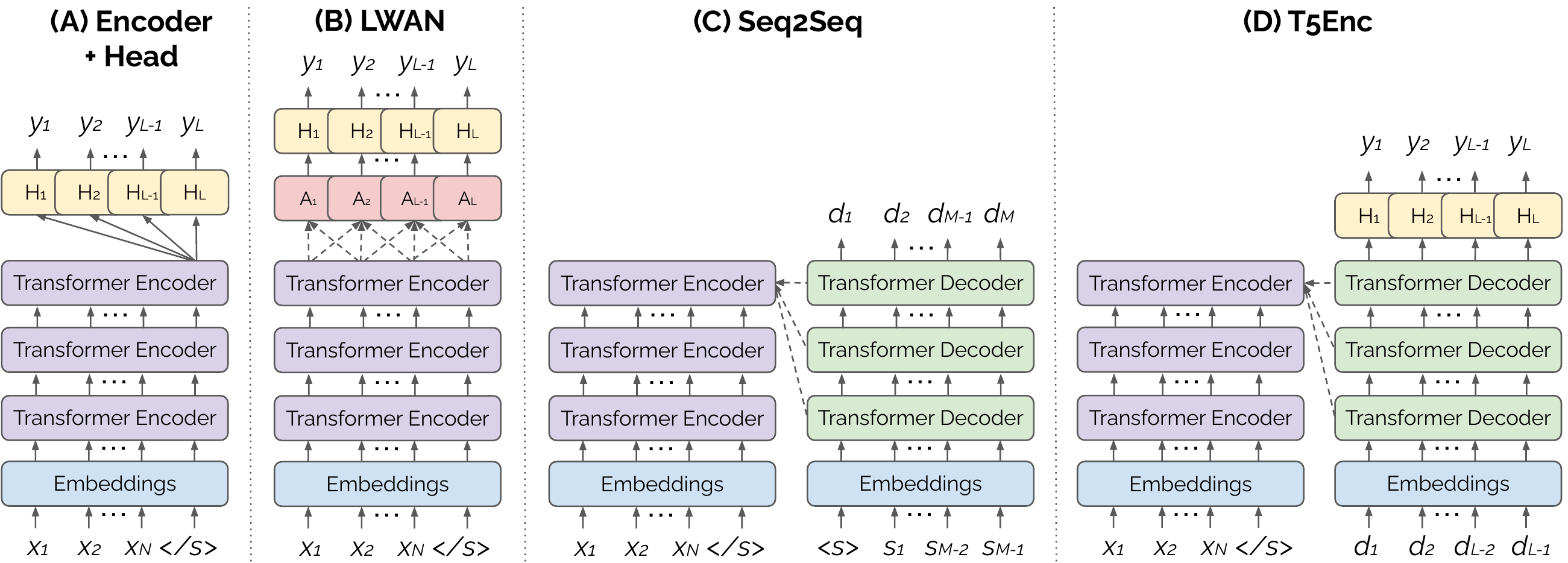}
    }
    \caption{Depiction of the four task-specific methods for multi-label classification: encoder-only (\emph{Encoder+Head}, \emph{LWAN}), and encoder-decoder (\emph{Seq2seq}, \emph{T5Enc}). \textit{x}: input tokens, \textit{y}: label predictions, \textit{d}: label descriptors, \textit{N}: input sequence length, \textit{L}: label set size, \textit{M}: number of labels for the given input.}
    \label{fig:models}
\end{figure*}

\section{Methods}
\label{sec:methods}

We experiment with four methods for multi-label classification, \emph{Encoder+Head}, \emph{LWAN}, \emph{Seq2Seq}, and \emph{T5Enc}, basing their implementation on the T5 model \cite{raffel-etal-2020}.
T5 is a transformer-based encoder-decoder model~\cite{Vaswani2017}, which encodes a string of input tokens and generates a string of output tokens. 

All methods discussed below use T5's encoder to represent input documents, a document being denoted as $[x_1, x_2, \dots, x_N]$, where $N$ is the document length in terms of T5 subword tokens. Some methods further use the model's decoder---we introduce decoder notation where needed.

\paragraph{Encoder+Head} In this case, we use only the encoder of T5 in the standard classification setting, as introduced by \citet{devlin-etal-2019-bert}. We feed the document to the encoder, and use the representation of the special \cls token as document representation ($d\in{\rm I\!R}^{dim}$). This representation is passed to $L$ standard classification heads, one per label.

\paragraph{LWAN} In this case, we use a Label-Wise Attention Network (LWAN) \cite{Mullenbach2018} on top of the T5 encoder, as done in \citet{chalkidis-etal-2020-empirical}. We feed the document to the encoder, and use one attention head per label to generate $L$ label-wise document representations $d_l\in{\rm I\!R}^{dim}$, i.e., $L$ weighted averages of the contextualized token representations. Intuitively, each head focuses on possibly different tokens of the document relevant to the corresponding label. LWAN employs $L$ linear layers ($o_l\in{\rm I\!R}^{dim\times1}$) each operating on a different label-wise document representation $d_l$, to produce $L$ scores (logits), one per label.

\paragraph{Seq2Seq} In this case, we use T5 for conditional generation, which is the standard form of use, since T5 was trained in an autoregressive fashion. The target labels are formatted as a sequence of label descriptors, separated by a comma and a space, and ordered alphabetically, e.g., `EU, finance'. We feed the document to the encoder and use the decoder to generate the tokenized output sequence, $[s_1, s_2, \dots, s_M]$. When we evaluate the trained model's performance in inference time, we split the generated sequences using comma as a delimiter, keeping only valid label descriptors, and treat them as a set (since their order does mot matter for the task). We consider different options for the label descriptors, discussed in Section~\ref{subsec:preliminary}.

\paragraph{T5Enc} In this case, we follow the work of \citet{liu-etal-2022-t5enc}, where they use T5 in a non-autoregressive fashion.\footnote{We keep the name T5Enc, as coined by the authors, for consistency, although the model actually uses both the encoder and the decoder of T5.} We feed the document to the encoder, and use the decoder in non-autoregressive fashion, where its inputs are fixed (pre-populated), i.e., we feed the decoder with single-token label descriptors, $[d_1, d_2, ..., d_L]$, where $L$ is the size of the full label set. We then use a binary classification head ($o_l\in{\rm I\!R}^{dim\times1}$) per decoder output representation to produce $L$ scores, one per label. This method can be seen as an advanced version of the LWAN method which builds label-wise representations ($d_l$) via attention. In this case, however, these representations are further co-attended (conditioned) via the standard decoder self-attention across many decoder layers.

\section{Datasets}

\label{sec:datasets}

We experiment with four datasets from the legal and biomedical domains, each with two different label granularities, i.e., label sets including more abstract or more specialized concepts.

\paragraph{UKLEX} United Kingdom (UK) legislation is publicly available as part of the United Kingdom's National Archives.\footnote{\url{https://www.legislation.gov.uk/}} Most of the laws have been categorized in thematic categories (e.g., health-care, finance, education, transportation, planing), which are stated in the document preamble and are used for archival indexing purposes. The UKLEX dataset \cite{chalkidis-sogaard-2022-improved} comprises 36.5k UK laws. The dataset is chronologically split in training (20k, 1975--2002), development (8k, 2002--2008), and test (8.5k, 2008--2018) sets. 

\paragraph{EURLEX} European Union (EU) legislation is published on the EUR-Lex website. All EU laws are annotated by EU's Publications Office with multiple concepts from EuroVoc, a thesaurus maintained by the Publications Office.\footnote{\url{http://eurovoc.europa.eu/}} 
EuroVoc has been used to index documents in systems of EU institutions. We use the English part of the dataset of \citet{chalkidis2021-multieurlex}, which comprises 65k EU laws (documents). The dataset is chronologically split in training (55k, 1958--2010), development (5k, 2010--2012), and test (5k, 2012--2016) sets. It supports four different label granularities. We use the 1st and 2nd level of the EuroVoc taxonomy.

\paragraph{BIOASQ} The BIOASQ (Task A) dataset consist of biomedical articles from PubMed,\footnote{\url{https://pubmed.ncbi.nlm.nih.gov}} annotated with concepts from the Medical Subject Headings (MeSH) taxonomy \cite{tsatsaronis-etal-2015-bioasq,bioasq2021}.\footnote{\url{https://www.nlm.nih.gov/mesh/}} MeSH is a hierarchically-organized vocabulary produced by the National Library of Medicine. The current version of MeSH contains more than 29k concepts referring to various aspects of the biomedical research (e.g., diseases, chemicals and drugs). It is primarily used for indexing, cataloging, and searching of biomedical and health-related information. We subsample 100k documents from the period 2000-2021 in the latest version (v.2022) of the dataset, and split those chronologically for training (80k, 1964--2015), development (10k, 2015--2018), and testing (10k, 2018--2020). We use the 1st and 2nd levels of the MeSH taxonomy.

\paragraph{MIMIC-III} The MIMIC-III dataset \cite{Johnson2017} contains approximately 50k discharge summaries from US hospitals. Each summary is annotated with one or more codes (labels) from the ICD-9 hierarchy, which has eight levels in total.\footnote{\url{www.who.int/classifications/icd/en/}}. The International Classification of Diseases, Ninth Revision (ICD-9) is the official system of assigning codes to diagnoses and procedures associated with hospital utilization in the United States. Documents in MIMIC-III have been anonymized to protect patient privacy, including chronological information (e.g., entry/discharge dates). Hence, it is not possible to split the data chronologically, so we split it randomly in train (30k), development (10k), and test (10k) sets. We use the 1st and 2nd level of the ICD-9 hierarchy.

All four datasets come with label descriptors, e.g. `Agriculture \& Food', `Immigration \& Citizenship' (UKLEX), and `Chemicals and Drugs', `Skin and Connective Tissue Diseases' (BIOASQ).\footnote{See Appendix~\ref{sec:dataset_descr} for label descriptors across all datasets.} More details about the datasets are provided in Table~\ref{tab:data_stats}. Notice that Level 2 label sets are considerably larger than Level 1 label sets, and that the number of label assignments per document do not grow proportionately from Level 1 to Level 2, which means Level 2 labels have less representation on average. 

\begin{table}[]
    \centering
    \resizebox{\columnwidth}{!}{
    \begin{tabular}{c|c|ccc|ccc}
    \toprule
        Dataset & Size & $|$L1$|$ & L/D & T/L&  $|$L2$|$ & L/D & T/L \\ 
        \midrule
        UKLEX & 36.5k & 18 & 1.2 & 2.1 & 69 & 1.5 & 1.7\\
        EURLEX & 65k & 21  & 3.2 & 2.4 & 127 & 4.5 & 2.9\\
        BIOASQ & 100k & 16 & 5.6 & 3.4 & 116 & 8.9 & 4.0 \\
        MIMIC-III & 50k & 19 & 6.0 & 7.8 & 184 & 10.1 & 8.4\\
    \bottomrule
    \end{tabular}}
    \caption{Summary of datasets in terms of size, number of labels on Level 1 ($|$L1$|$) and 2 ($|$L2$|$), average number of gold labels per document (L/D), and average number of tokens per label (T/L) in the T5 vocabulary.}
    \label{tab:data_stats}
    \vspace{-2mm}
\end{table}

\section{Experiments}

\subsection{Experimental Setup}

We use the original checkpoints of T5 released by \citet{{raffel-etal-2020}} from the Hugging Face Hub.\footnote{\url{https://huggingface.co/t5-base}} Following \citeauthor{raffel-etal-2020}, for all four methods we use the Adafactor optimizer~\cite{shazeer-etal-2018} with a fixed learning rate of 1e-4 after warm-up for one epoch.\footnote{In preliminary experiments, we also considered the widely used AdamW optimizer~\cite{Loshchilov2017}, which led to lower performance in most cases.} Seq2Seq models are trained with teacher forcing.  We report results in terms of micro-F1 (\microf), and macro-F1 (\macrof) scores, the former more indicative of performance on well-represented labels, the latter, of performance on rare labels. When fine-tuning models, we use early stopping based on validation micro-F1 scores. We run each experiment with 4 seeds, and report the mean and standard deviations across runs.

\subsection{Preliminary Experiments}\label{subsec:preliminary}

We conduct a series of preliminary experiments to identify the most promising setting for the examined methods. All results reported here are on the development split of respective datasets.

\begin{table}[h]
    \centering
    \resizebox{\columnwidth}{!}{
    \begin{tabular}{l|cc|cc}
    \toprule
    \multirow{2}{*}{\bf No. Heads} &  \multicolumn{2}{c|}{\bf UKLEX (L1)} & \multicolumn{2}{c}{\bf EURLEX (L2)} \\ 
     & \microf & \macrof & \microf & \macrof \\
     \midrule
         N=1 & \bf 83.3 $\pm$ \small{0.2} & \bf 79.3 $\pm$ \small{0.7} & \bf 76.3 $\pm$ \small{0.3} & \bf 55.5 $\pm$ \small{0.8} \\
         N=4 & 82.8 $\pm$ \small{0.3} & 78.1 $\pm$ \small{0.7} & 75.1 $\pm$ \small{0.1} & 51.7 $\pm$ \small{2.1} \\
         N=6 & 83.2 $\pm$ \small{0.3} & \bf 79.3 $\pm$ \small{0.5} & 75.1 $\pm$ \small{0.3} & 54.1 $\pm$ \small{0.6} \\
        N=12 & 83.0 $\pm$ \small{0.4} & 78.8 $\pm$ \small{1.4} & 75.2 $\pm$ \small{0.3} & 53.0 $\pm$ \small{1.2} \\
     \bottomrule
    \end{tabular}
    }
    \caption{Number of attentions heads for LWAN.}
    \label{tab:lwan_heads}
\end{table}

\paragraph{LWAN -- Number of attention heads}
Previous work which employed the LWAN approach always used a single attention head in the label-wise attention mechanism.
Here, we experiments with $N\in[1,4,6,12]$. In Table~\ref{tab:lwan_heads}, we reports results on two datasets, UKLEX (L1) with 18 labels, and EURLEX (L2) with 127 labels. We observe that in the case of UKLEX (L1) increasing the number of attention heads does not improve results, while in the case of EURLEX (L2) it harms performance. It appears that the added expressivity from multi-head attention is either not needed, or it is not easily utilized, since it adds more randomly initialized parameters which have to be learned from scratch. In subsequent experiments, we thus use the standard single-head attention mechanism.

\begin{table}[h]
    \centering
    \resizebox{\columnwidth}{!}{
    \begin{tabular}{l|cc|cc}
    \toprule
    \multirow{2}{*}{\bf Label} &  \multicolumn{2}{c|}{\bf UKLEX (L1)} & \multicolumn{2}{c}{\bf MIMIC (L1)} \\ 
     & \microf & \macrof & \microf & \macrof \\
     \midrule
  Original & 84.2 $\pm$ \small{0.0} & \bf 81.6 $\pm$ \small{0.2} & 73.2 $\pm$ \small{0.0} & \bf 70.2 $\pm$ \small{0.2} \\
  Simplified & \bf 84.8 $\pm$ \small{0.2} & 78.7 $\pm$ \small{0.3} & 73.1 $\pm$ \small{0.1} & 70.1 $\pm$ \small{0.1} \\
   Numbers & 83.8 $\pm$ \small{0.2} & 80.2 $\pm$ \small{0.7} & \bf 73.3 $\pm$ \small{0.1} & 69.7 $\pm$ \small{0.2} \\
 \bottomrule
    \end{tabular}
    }
    \caption{Form of label descriptors for Seq2Seq.}
    \label{tab:label_descs}
\end{table}

\paragraph{Seq2Seq -- Form of Label Descriptors}

We consider three alternative forms of label descriptors:\vspace{1mm}

\begin{enumerate}[label=(\alph*)]
    \item the \emph{original} label descriptors, which may include complex multi-word expressions, e.g., `Anthropology, Education, Sociology, and Social Phenomena'
    \item \emph{simplified} versions of the original label descriptors, manually curated to consist of single-token expressions (as per the T5 vocabulary), e.g., `Anthropology' for the example above
    \item \emph{numbers} arbitrarily assigned to labels, e.g. `1'. In Table~\ref{tab:label_descs}, we present results on two datasets, UKLEX (L1), where the original label descriptors are mostly single-word expressions that map onto T5 sub-word tokens, and MIMIC (L1), where the original label descriptors are multi-word expressions which are further tokenized into subwords
\end{enumerate}

\noindent We observe mixed rankings between the three forms of label descriptors across different metrics and datasets, with slight advantage for a lexical form over the arbitrary numerical one. This is in line with the intuition that the semantics of the label descriptors contribute to the learning of the task. 
In subsequent experiments, we use the original label descriptors across all datasets.

\begin{table}[h]
    \centering
    \resizebox{\linewidth}{!}{
    \begin{tabular}{l|cc|cc}
    \toprule
    \multirow{2}{*}{\bf Decoding} &  \multicolumn{2}{c|}{\bf UKLEX (L1)} & \multicolumn{2}{c}{\bf MIMIC (L1)} \\ 
     & \microf & \macrof & \microf & \macrof \\ 
     \midrule
Greedy & \bf 84.3 $\pm$ \small{0.0} & \bf 81.6 $\pm$ \small{0.2} &  72.9 $\pm$ \small{0.2} & 69.4 $\pm$ \small{0.4} \\
Beam & 84.2 $\pm$ \small{0.0} & \bf 81.6 $\pm$ \small{0.2} & \bf 73.2 $\pm$ \small{0.1} & \bf 70.3 $\pm$ \small{0.2} \\ 
 \bottomrule
    \end{tabular}
    }
    \caption{Greedy decoding vs. beam search for Seq2Seq.}
    \label{tab:decoding}
    \vspace{-.5mm}

\end{table}

\paragraph{Seq2Seq -- Greedy Decoding vs. Beam Search}

\citet{raffel-etal-2020} suggested using greedy decoding for single-label classification tasks but also found beam search decoding (N=4) to work better for tasks with long output sequences, as is the case in multi-label classification.
In Table~\ref{tab:decoding}, we compare the two decoding strategies on UKLEX (L1) and MIMIC (L1). We find that the choice of decoding strategy has little effect on performance, likely because the output space in these tasks is constrained to a fixed set of valid labels, in a single permissible (alphabetical) order. In subsequent experiments, we use beam search (N=4), as it performs slightly better on average.

\begin{table}[h]
    \centering
    \resizebox{\columnwidth}{!}{
    \begin{tabular}{l|cc|cc}
    \toprule
    \multirow{2}{*}{\bf Label} &  \multicolumn{2}{c|}{\bf UKLEX (L1)} & \multicolumn{2}{c}{\bf MIMIC (L1)} \\ 
     & \microf & \macrof & \microf & \macrof \\
     \midrule
 Simplified & \bf 84.8 $\pm$ \small{0.2} & 81.9 $\pm$ \small{0.5} & \bf 73.6 $\pm$ \small{0.2} & \bf 69.2 $\pm$ \small{1.5}     \\
 Pseudo &  \bf 84.8 $\pm$ \small{0.1} & \bf 82.3 $\pm$ \small{0.2} & 73.2 $\pm$ \small{0.1} & 67.7 $\pm$ \small{1.9} \\
 \bottomrule
    \end{tabular}
    }
    \caption{Form of label descriptors for T5Enc.}
    \label{tab:label_descs_t5}
    \vspace{-.5mm}
\end{table}

\begin{table*}[h]
    \centering
    \resizebox{\textwidth}{!}{
    \begin{tabular}{l|cc|cc|cc|cc|cc}
    \toprule
         \multirow{2}{*}{\bf Method} &  \multicolumn{2}{c|}{\bf UKLEX (L1)} & \multicolumn{2}{c|}{\bf EURLEX (L1)} & \multicolumn{2}{c|}{\bf BIOASQ (L1)} & \multicolumn{2}{c}{\bf MIMIC (L1)} & \multicolumn{2}{c}{\bf Average}\\
         & \microf & \macrof & \microf & \macrof & \microf & \macrof & \microf & \macrof & \microf & \macrof \\
         \midrule
   Enc+Head & \bf 80.8 $\pm$ \small{0.5} & \bf 77.2 $\pm$ \small{0.4} & 78.9 $\pm$ \small{0.4} & 67.9 $\pm$ \small{1.1} & 86.4 $\pm$ \small{0.0} & 76.8 $\pm$ \small{0.1} & 72.2 $\pm$ \small{0.2} & 66.3 $\pm$ \small{0.7} & 79.6 & 72.1\\
      LWAN & 80.4 $\pm$ \small{0.3} & 76.6 $\pm$ \small{0.5} & 79.6 $\pm$ \small{0.4} & 68.4 $\pm$ \small{0.7} & 86.3 $\pm$ \small{0.1} & 77.2 $\pm$ \small{0.2} & 72.3 $\pm$ \small{0.3} & 66.8 $\pm$ \small{0.8} & 79.7 & 72.3\\
  \midrule
  Seq2Seq & 79.6 $\pm$ \small{0.6} & 76.4 $\pm$ \small{0.6} & 78.8 $\pm$ \small{0.2} & 69.1 $\pm$ \small{0.3} & 86.0 $\pm$ \small{0.1} & 77.8 $\pm$ \small{0.2} & 72.9 $\pm$ \small{0.1} & \bf 69.7 $\pm$ \small{0.2} & 79.3 & 73.3\\
  T5Enc & \bf 80.8 $\pm$ \small{0.4} & 77.1 $\pm$ \small{0.5} & \bf 80.0 $\pm$ \small{0.3} & \bf 70.5 $\pm$ \small{0.4} & \bf 86.6 $\pm$ \small{0.0} & \bf 77.9 $\pm$ \small{0.4} & \bf 73.4 $\pm$ \small{0.3} & 68.8 $\pm$ \small{1.4} & \bf 80.2 & \bf 73.6\\
\midrule
\midrule
\multirow{2}{*}{\bf Method} &  \multicolumn{2}{c|}{\bf UKLEX (L2)} & \multicolumn{2}{c|}{\bf EURLEX (L2)} & \multicolumn{2}{c|}{\bf BIOASQ (L2)} & \multicolumn{2}{c}{\bf MIMIC (L2)} & \multicolumn{2}{c}{\bf Average}\\
         & \microf & \macrof & \microf & \macrof & \microf & \macrof & \microf & \macrof & \microf & \macrof \\
         \midrule
  Enc+Head & 75.9 $\pm$ \small{0.5} & 64.9 $\pm$ \small{0.5} & 70.3 $\pm$ \small{0.2} & 48.2 $\pm$ \small{1.2} & 73.1 $\pm$ \small{0.0} & 60.1 $\pm$ \small{0.8} & 56.7 $\pm$ \small{0.6} & 22.3 $\pm$ \small{1.2} & 69.0 & 48.9\\
  LWAN & \bf 76.6 $\pm$ \small{0.2} & 65.0 $\pm$ \small{0.8} & 70.3 $\pm$ \small{0.3} & 49.0 $\pm$ \small{0.7} & 73.0 $\pm$ \small{0.1} & 59.7 $\pm$ \small{0.9} & 57.2 $\pm$ \small{0.4} & 24.2 $\pm$ \small{0.3} & 69.3 & 49.5 \\
      \midrule
 Seq2Seq & 75.3 $\pm$ \small{0.2} & 65.8 $\pm$ \small{0.4} & 70.6 $\pm$ \small{0.3} & 51.8 $\pm$ \small{1.0} & 73.8 $\pm$ \small{0.1} & 63.8 $\pm$ \small{0.1} & 57.4 $\pm$ \small{0.2} & \bf 31.2 $\pm$ \small{1.7} & 69.3 & 53.2 \\
 T5Enc & 76.5 $\pm$ \small{0.3} & \bf 66.8 $\pm$ \small{0.9} & \bf 72.0 $\pm$ \small{0.2} & \bf 53.2 $\pm$ \small{1.4} & \bf 75.1 $\pm$ \small{0.1} & \bf 66.0 $\pm$ \small{0.1} & \bf 60.5 $\pm$ \small{0.1} & 31.1 $\pm$ \small{0.9} & \bf 71.0 & \bf 54.3 \\
 \bottomrule
    \end{tabular}
    }
    \caption{Test results for encoder-only methods (Encoder+Head and LWAN) and encoder-decoder methods (Seq2Seq and T5Enc) trained from T5-Base.}
    \label{tab:overall_t5}
\end{table*}

\paragraph{T5Enc -- Form of Label Descriptors}

We compare two forms of label tokens, lexical (using simplified descriptors, as they have to be single tokens), and pseudo descriptors, where we introduce special tokens to the vocabulary of T5 (e.g., \pseudolabel). Results on UKLEX (L1) and MIMIC (L1) are presented in Table~\ref{tab:label_descs_t5}. We observe that results are comparable for UKLEX, while simplified label descriptors perform slightly better for MIMIC. In subsequent experiments, we thus use simplified label descriptors for Level 1 datasets. For Level 2 datasets, we use pseudo labels, since we cannot manually curate simplified descriptors for hundreds of labels.

\begin{table}[h]
    \centering
    \resizebox{\columnwidth}{!}{
    \begin{tabular}{l|cc|cc}
    \toprule
    \multirow{2}{*}{\bf Encoder} &  \multicolumn{2}{c|}{\bf UKLEX (L1)} & \multicolumn{2}{c}{\bf BIOASQ (L2)} \\ 
     & \microf & \macrof & \microf & \macrof \\
     \midrule
BERT & \textbf{84.4} $\pm$ \small{0.3} & \textbf{81.3} $\pm$ \small{0.9} & 71.7 $\pm$ \small{0.0} & 59.1 $\pm$ \small{0.0} \\
RoBERTa & 84.3 $\pm$ \small{0.6} & 81.1 $\pm$ \small{1.1} & 73.0 $\pm$ \small{0.0} & 59.8 $\pm$ \small{0.0} \\
T5 & 84.3 $\pm$ \small{0.3} & 80.7 $\pm$ \small{0.8} & \textbf{73.2} $\pm$ \small{0.1} & \textbf{60.8} $\pm$ \small{0.8} \\
 \bottomrule
    \end{tabular}
    }
    \caption{Encoder-only pre-trained models vs. T5's encoder in Encoder+Head classification setups.}
    \label{tab:bert_models}
\end{table}

\paragraph{Encoder-only Models}

Comparing encoder-only to encoder-decoder methods fro multi-label text classification in a fair manner is non-trivial since inherently encoder-only pre-trained models like BERT~\cite{devlin-etal-2019-bert}, and RoBERTa~\cite{liu-2019-roberta} are trained on different data and with a different objective than the encoder-decoder model T5. Using T5's encoder for encoder-only methods circumvenes this problem but introduces another concern: that this encoder was trained in an encoder-decoder architecture and may thus be handicapped in comparison to encoders trained in an encoder-only architecture. 

In Table~\ref{tab:bert_models}, we present development results on UKLEX (L1) and BIOASQ (L2) for encoder-only classifiers trained from BERT, RoBERTa and T5's encoder.\footnote{We use the prepended \texttt{[CLS]} token representation for BERT and RoBERTa.} We observe mixed results with BERT performing best on UKLEX (L1) and T5 performing best on EURLEX (L2), with absolute differences between the three models being relatively small and on average between the two datasets, favouring T5. We thus conclude that T5's encoder makes for a fair and strong encoder-only baseline and use it in subsequent experiments.

\subsection{Main Results}
\label{sec:experiments}

In Table~\ref{tab:overall_t5}, we present test results for all methods trained from T5-Base.\footnote{We present development results in Table~\ref{tab:overall_t5_dev} in Appendix~\ref{sec:additional_results} for completeness.} The overall best performing approach is T5Enc, followed by Seq2Seq, LWAN and then Encoder+Head. The trend is thus for encoder-decoder approaches (T5Enc and Seq2Seq) to outperform encoder-only approaches (LWAN and then Encoder+Head), which use just half the model parameters. This result corroborates and considerably substantiates the observations of \citet{liu-etal-2022-t5enc}. We gain further insights through a breakdown by metric and label granularity.

The advantage of encoder-decoder methods can be especially seen across macro-F1 scores, where both T5Enc and Seq2Seq outperform encoder-only approaches almost categorically (the one exception being UKLEX (L1)). This indicates that encoder-decoder approaches are particularly good at assigning less frequent labels, which is a key challenge in multi-label classification. This reading of the results is further reinforced by the observation that the performance gap increases from Level 1 datasets, which contain a smaller number of labels, to Level 2 datasets, which contain more and thus on average less frequent labels. The most striking performance gap we observe measures 7 p.p. between LWAN and Seq2Seq on MIMIC (L2).

Between the two encoder-decoder approaches, we see that the non-autoregressive use of the T5 decoder is more effective (T5Enc) than the conditional generation of labels (Seq2Seq), the gap between the two methods growing from Level 1 to Level 2 datasets. In the case of T5Enc, the decoder serves to build representations for all labels relevant to a dataset and in this sense defines and constraints the output space for the task. Meanwhile, in the Seq2Seq approach the model has to learn the constraints on the output space during training, and as such it is likely more prone to errors.

These main results give us a general idea of how the different approaches compare, indicating clearly that encoder-decoder approaches are superior. In subsequent sections we explore the source of performance and the limitations of encoder-decoder approaches further. 

\begin{figure*}[ht]
\begin{subfigure}{.254\textwidth}
  \centering
  \includegraphics[width=\linewidth]{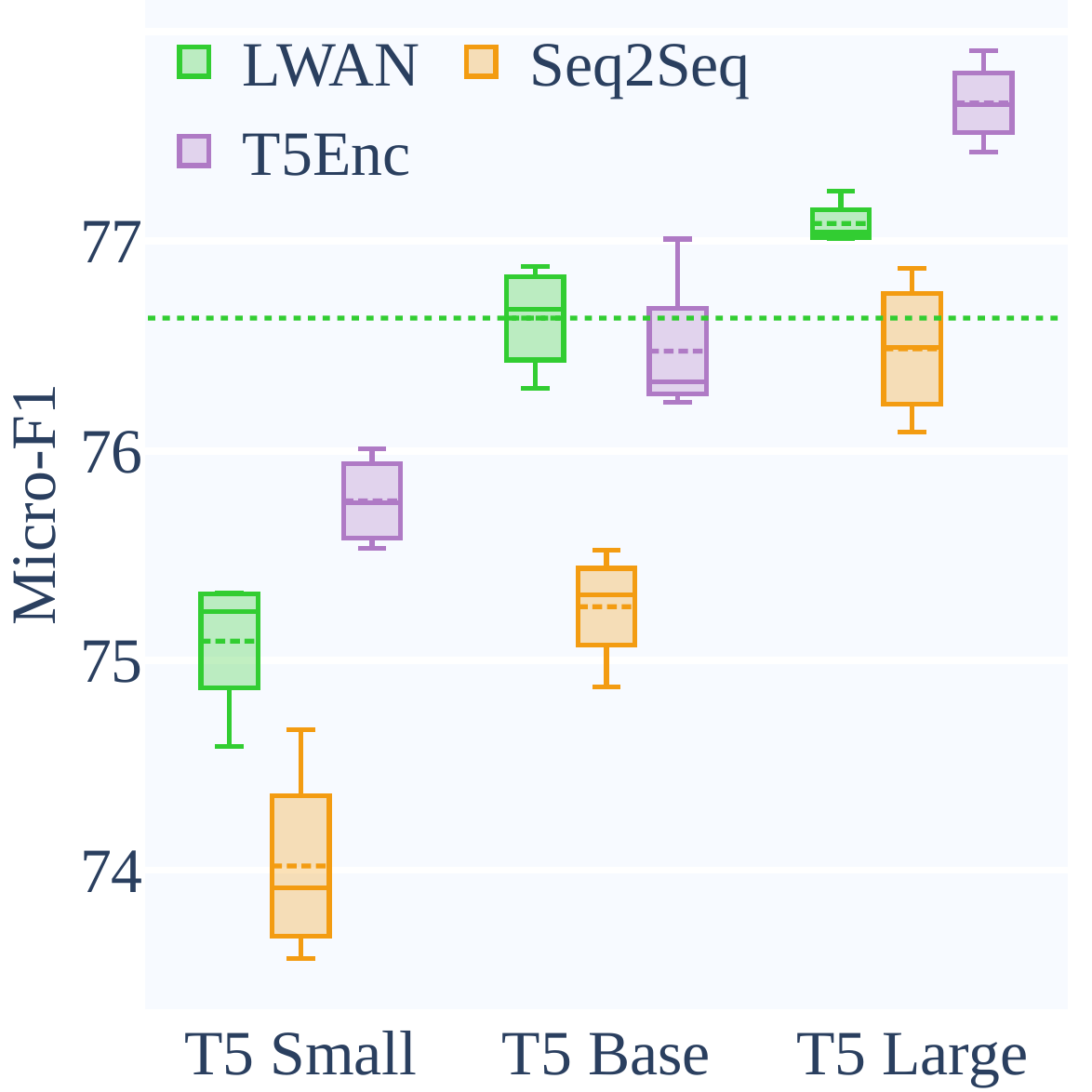}  
  \caption{UKLEX}
  \label{fig:sub-first}
\end{subfigure}
\begin{subfigure}{.2355\textwidth}
  \centering
  \includegraphics[width=\linewidth]{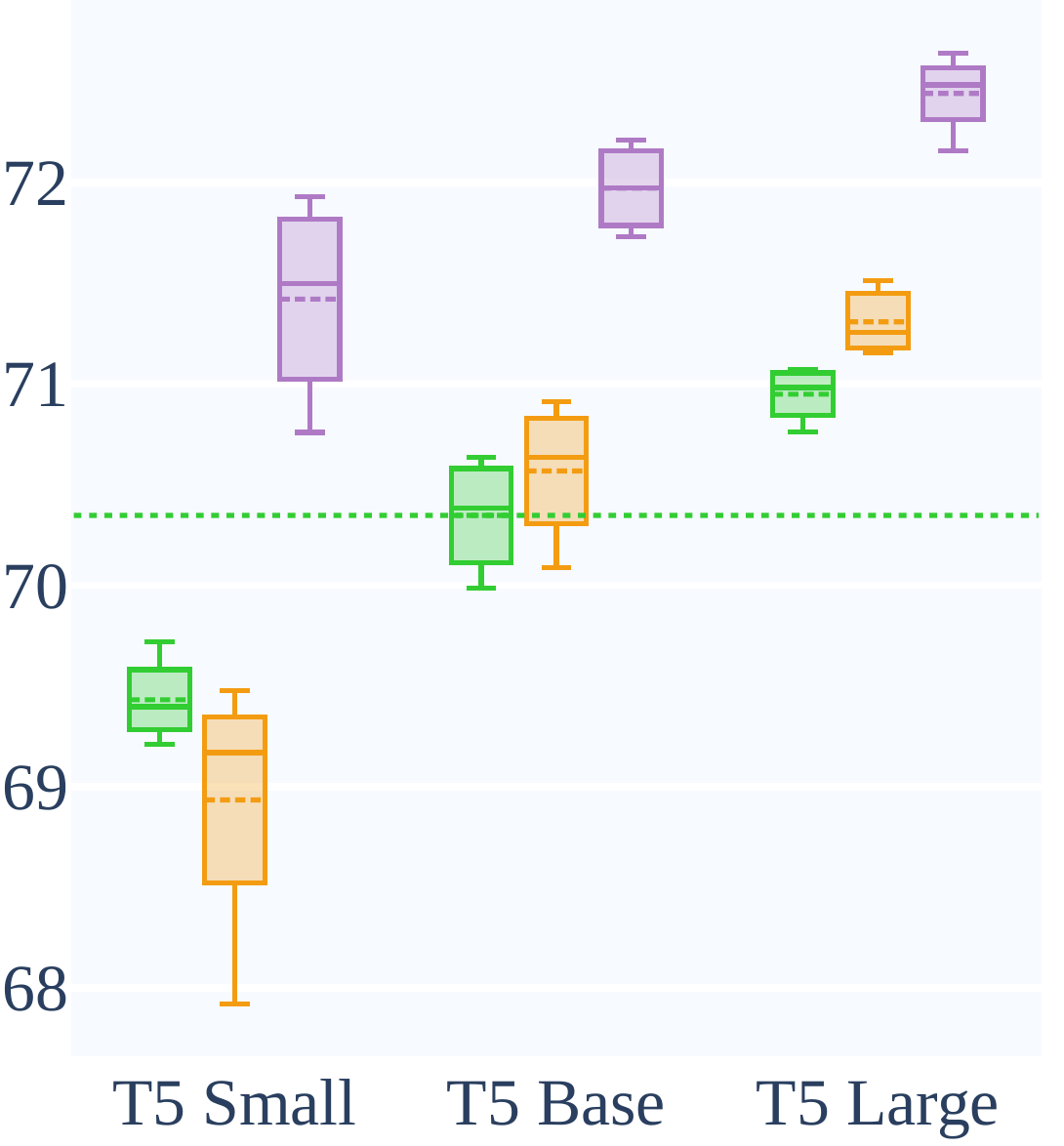}  
  \caption{EURLEX}
  \label{fig:sub-second}
\end{subfigure}
\begin{subfigure}{.2355\textwidth}
  \centering
  \includegraphics[width=\linewidth]{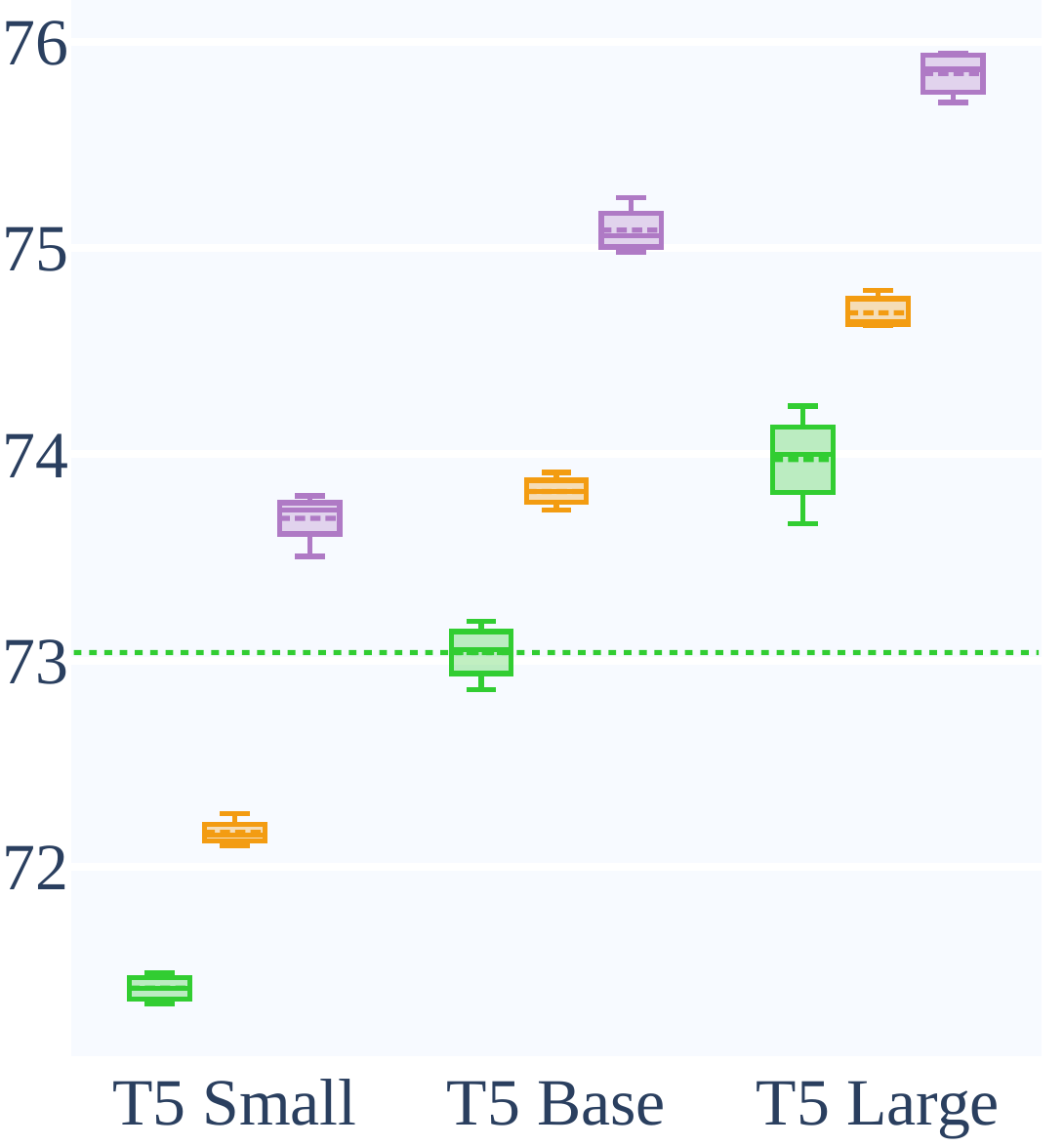}  
  \caption{BIOASQ}
  \label{fig:sub-third}
\end{subfigure}
\begin{subfigure}{.2355\textwidth}
  \centering
  \includegraphics[width=\linewidth]{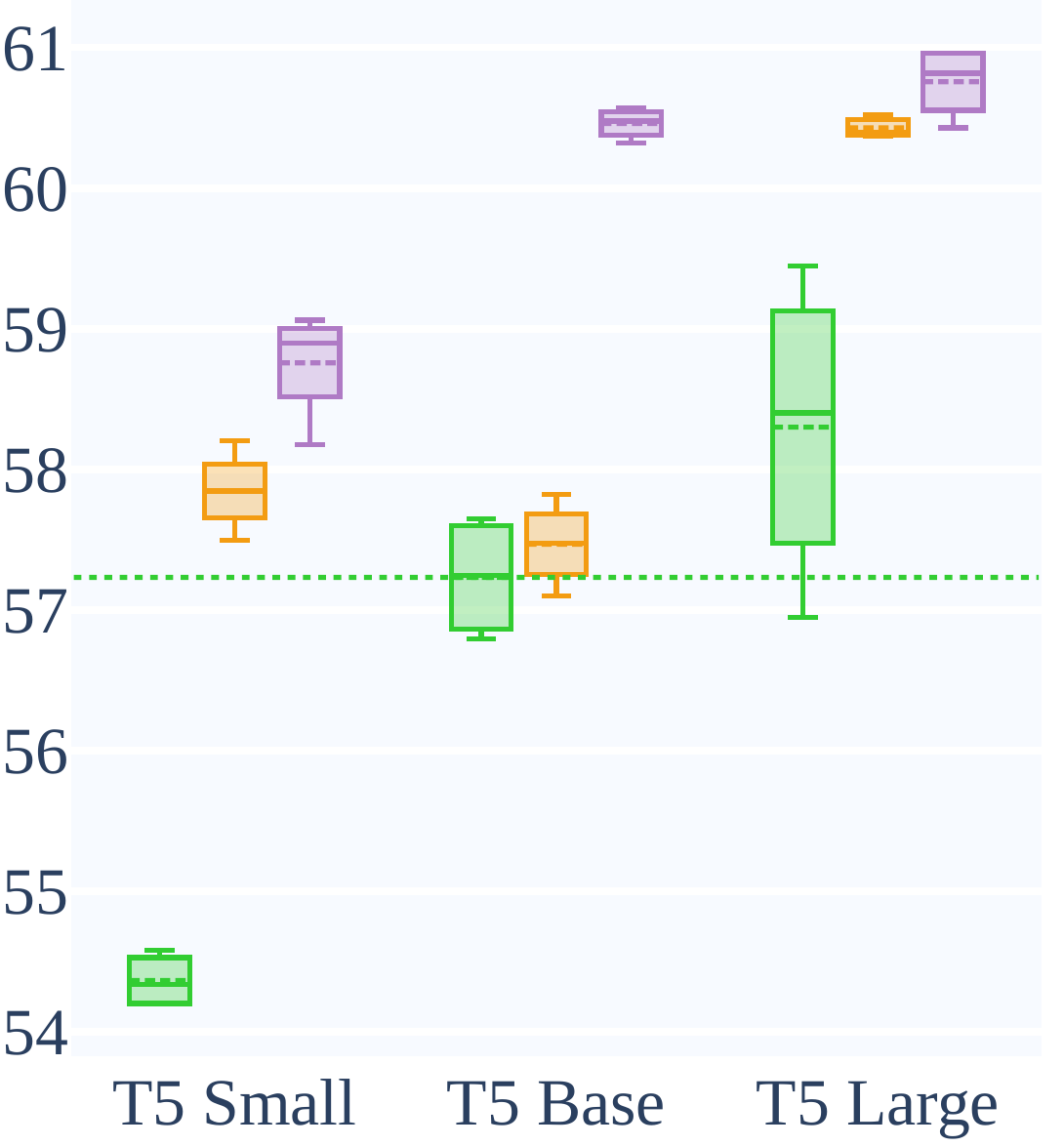}  
  \caption{MIMIC}
  \label{fig:sub-fourth}
\end{subfigure}
\caption{Performance of the three strongest classification methods (\emph{LWAN},  \emph{Seq2Seq}, \emph{T5Enc}) across three model sizes in terms of micro-F1 score. Dashed lines inside the boxes represent the mean performance across four seeds.}
\label{fig:model_size}
\end{figure*}

\subsection{Model Capacity}

One possible explanation for the stronger performance of encoder-decoder methods is that they operate with twice as many parameters as encoder-only methods. Here, we test whether this alone is the source of their improved performance, by training models from different T5 models: small, base and large.\footnote{T5-Small has 12 layers of d=512, T5-Base has 24 layers of d=768, T5-Large has 48 layers of d=1024, where half of the layers are in the encoder and half in the decoder.} Since we previously saw that trends in results are similar across L1 and L2 datasets, and more pronounced in the latter, we carry out this set of experiments on L2 datasets only. We include the stronger performing encoder-only approach, LWAN, as well as both encoder-decoder approaches. Results on the micro-F1 metric are presented in Figure~\ref{fig:model_size}, and on the macro-F1 metric in Figure~\ref{fig:model_size2} in Appendix~\ref{sec:additional_results}.\footnote{All results are also presented in Table~\ref{tab:overall_t5_small_large} in Appendix~\ref{sec:additional_results}.} 

Firstly, we note that T5Enc consistently outperforms the other approaches across different model sizes, in line with earlier findings (see Table~\ref{tab:overall_t5}). We also see that all methods appear to scale, with steady improvements in performance observed across increasing model sizes.

Comparing models of similar size (i.e., models with the same number of layers), we gain a more precise idea of how methods compare. Here, T5Enc still proves to be the superior approach, with T5Enc-Small outperforming LWAN-Base on 3 out of 4 datasets (UKLEX being the exception), and similarly T5Enc-Base outperforming LWAN-Large on 3 out of 4 datasets. 
Notice that in these comparisons, the T5Enc variants are even at a disadvantage, having the same number of layers as the LWAN variants, but lower dimensionality. Seq2Seq models, on the other hand, underperform similarly-sized LWAN models on most comparisons in terms of micro-F1, which indicates that this approach is overall less suitable for the task.\footnote{See Appendix~\ref{sec:additional_results} for a discussion of macro-F1 results.} 

\subsection{Ablations on T5Enc Decoder}

Here, we analyse the contribution of different aspects of the T5Enc decoder through ablations on the decoder's depth, width and self-attention. 

\paragraph{Decoder Depth}

We train T5Enc models with a varying number of decoder layers. We experiments with $N\in[1,4,6,12]$. In Table~\ref{tab:dec_layers}, we report results on two datasets, UKLEX (L1) and EURLEX (L2). We observe that larger depth in the decoder contributes to performance, with the full set of decoder layers (12) performing best.

\begin{table}[h]
    \centering
    \resizebox{\columnwidth}{!}{
    \begin{tabular}{l|cccc}
    \toprule
    \multirow{2}{*}{\bf Layers} &  \multicolumn{2}{c}{\bf UKLEX (L1)} & \multicolumn{2}{c}{\bf EURLEX (L2)} \\ 
     & \microf & \macrof & \microf & \macrof \\
     \midrule
         N=1 & 84.6 $\pm$ \small{0.1} & 81.9 $\pm$ \small{0.1} & 76.6 $\pm$ \small{0.1} & 56.9 $\pm$ \small{0.1} \\
         N=4 &  84.7 $\pm$ \small{0.1} & 81.8 $\pm$ \small{0.1} & 76.9 $\pm$ \small{0.1} & 58.1 $\pm$ \small{1.1} \\
         N=6 & \bf 84.8 $\pm$ \small{0.1} & \bf 82.2 $\pm$ \small{0.1} & 77.0 $\pm$ \small{0.1} & 58.4 $\pm$ \small{1.3} \\
        N=12 & \bf 84.8 $\pm$ \small{0.2} & 81.9 $\pm$ \small{0.5} & \bf 77.1 $\pm$ \small{0.1} & \bf 58.8 $\pm$ \small{1.4}  \\
    \bottomrule
    \end{tabular}
    }
    \caption{Development results for different numbers of decoder layers in T5Enc.}
    \label{tab:dec_layers}
\end{table}

\begin{table*}[t]
    \centering
    \resizebox{\textwidth}{!}{
    \begin{tabular}{l|cc|cc|cc|cc}
    \toprule
\multirow{2}{*}{\bf Method} &  \multicolumn{2}{c|}{\bf UKLEX (L2)} & \multicolumn{2}{c|}{\bf EURLEX (L2)} & \multicolumn{2}{c|}{\bf BIOASQ (L2)} & \multicolumn{2}{c}{\bf MIMIC (L2)} \\
         & \microf & \macrof & \microf & \macrof & \microf & \macrof & \microf & \macrof \\
         \midrule
Encoder+Head & 81.9 $\pm$ \small{0.6} & 72.9 $\pm$ \small{1.3} & 76.2 $\pm$ \small{0.2} & 54.0 $\pm$ \small{1.4} & 73.2 $\pm$ \small{0.1} & 60.8 $\pm$ \small{0.8} & 56.7 $\pm$ \small{0.7} & 22.3 $\pm$ \small{1.2} \\
\midrule
Single-step T5Enc & \bf 82.6 $\pm$ \small{0.1} & \bf 74.4 $\pm$ \small{0.8} & 76.7 $\pm$ \small{0.2} & 55.8 $\pm$ \small{1.4} & 73.5 $\pm$ \small{0.3} & 61.8 $\pm$ \small{1.1} & 58.3 $\pm$ \small{0.5} & 25.8 $\pm$ \small{0.9} \\
T5Enc & 82.4 $\pm$ \small{0.4} & 74.2 $\pm$ \small{1.0} & \bf 77.1 $\pm$ \small{0.1} & \bf 58.8 $\pm$ \small{1.4} & 75.1 $\pm$ \small{0.0} & \bf 66.3 $\pm$ \small{0.1} & \bf 60.6 $\pm$ \small{0.1} & 31.1 $\pm$ \small{1.0} \\
- No attention & 81.9 $\pm$ \small{0.1} & 73.0 $\pm$ \small{0.5} & 76.8 $\pm$ \small{0.1} & 57.6 $\pm$ \small{0.8} & 74.3 $\pm$ \small{0.1} & 64.3 $\pm$ \small{0.3} & 58.6 $\pm$ \small{0.3} & 27.4 $\pm$ \small{1.6} \\
- Full attention & 82.3 $\pm$ \small{0.2} & 74.1 $\pm$ \small{0.8} & \bf 77.1 $\pm$ \small{0.2} & 58.7 $\pm$ \small{0.8} & \bf 75.2 $\pm$ \small{0.0} & 66.1 $\pm$ \small{0.0} & \bf 60.6 $\pm$ \small{0.2} & \bf 31.6 $\pm$ \small{0.7} \\
 \bottomrule
    \end{tabular}
    }
    \caption{Ablations of T5Enc decoder. Single-step T5Enc builds a single output representation instead of $L$ label-wise representations. Attention ablations disable (No) or fully enable (Full) the self-attention in the decoder.}
    \label{tab:ablations}
\end{table*}

\paragraph{Decoder Width}
In this ablation, we are interested to establish the importance of label-wise representations being built in the decoder as opposed to using it to create a single output representation shared across the classification heads. 
To this end, we feed the decoder with a single token ID, e.g., the ID of token \emph{`label'}, and then pass its output representation ($d\in{\rm I\!R}^{dim}$) to a set of standard classification heads to produce $L$ scores (logits), similar to the Encoder+Head method. This method can be seen as an advanced version of the Encoder+Head method that utilizes the decoder via cross-attention. 

Results for Level 2 datasets are shown in Table~\ref{tab:ablations} under Single-step T5Enc (Level 1 results are shown in Table~\ref{tab:overall_t5_dev} in the Appendix). In comparison to the Encoder+Head baseline, Single-step T5Enc is superior across the board, likely because of the added number of parameters available to the model. Compared to the standard T5Enc approach, Single-step T5Enc works slightly better for UKLEX but on all other datasets it underperforms by a large gap. We observe the same pattern for L1 results in Table~\ref{tab:overall_t5_dev} and thus conclude that the additional computational power of label-wise processing is important for the good overall performance of T5Enc.

\paragraph{Attention Scheme}
The labels in multi-label classification are known to exhibit certain dependencies \cite{tenenboim2009multi, BOGATINOVSKI2022117215}. We measure the pair-wise dependency between labels in the four datasets included in this study, using Fisher's exact test.\footnote{The test determines whether the observed distribution of one variable is likely to be random given the observed distribution of another variable and vice-versa.} In Table~\ref{tab:fisher}, we report the percentage of label pairs in Level 2 label sets for which a significant association ($p<.001$) was discovered (see Appendix~\ref{sec:additional_results} for Level 1 results).
Based on the observed non-trivial rates of inter-label dependency, we hypothesize that self-attention in the T5 decoder is of key importance to the performance of T5Enc. 

\begin{table}[h]
    \centering
    \resizebox{0.97\linewidth}{!}{
    \begin{tabular}{l|cccc}
    \toprule
         \bf Level & \textbf{\textbf{\textbf{\textbf{UKLEX}}}} & \textbf{\textbf{\textbf{EURLEX}}} & \textbf{\textbf{BIOASQ}} & \textbf{MIMIC} \textbf{\textbf{}}\\
         \midrule
         L2 & 39.5 & 39.7 & 71.2 & 21.3 \\
    
    \bottomrule
    \end{tabular}}
    \caption{Percentage of Level 2 label pairs with significant association according to Fisher's exact test.}
    \label{tab:fisher}%
\end{table}

The decoder in T5 models uses \emph{causal} attention, wherein decoder inputs can only attend to the left context. We measure the contribution of this system component by ablating it, i.e. training T5Enc models with no self-attention. In Table~\ref{tab:ablations}, we report results on Level 2 datasets under \textit{No attention} (see  Table~\ref{tab:overall_t5_dev} in Appendix~\ref{sec:additional_results} for Level 1 results). We observe that without self-attention, performance suffers considerably for all datasets, most notably so in terms of macro-F1 on MIMIC ($\Delta=3.7$). This result indicates that self-attention indeed has a key role, although its contribution does not prove to be proportional to the rate of significant pair-wise associations in the data (Table~\ref{tab:fisher})---this may be due to higher-order label dependencies taking precedence over pair-wise ones.

Having confirmed the importance of modeling label dependency above, we next consider whether we can achieve even better performance with bidirectional (rather than causal) attention in the T5 decoder. In Table~\ref{tab:ablations} \textit{Full attention}, we see that the contribution of bidirectional attention is negligible. Assuming that the model is able to adjust to the new attention scheme during the fine-tuning process, we take these results to indicate that modeling label dependency in just one direction is sufficient. Indeed, Fisher's exact test measures two-way association, disregarding the direction of the dependency.

\subsection{Errors in Seq2Seq Models}\label{subsec:errors_seq2seq}

The Seq2Seq approach similarly can model label dependency through self-attention and can even condition the prediction of labels on one another (in an autoregressive fashion), an ability which none of the other approaches included in this study posses. Yet, we find empirically that it underperforms T5Enc. Here, we investigate whether this finding can be explained in terms of the unconstrained output space in Seq2Seq models. Specifically, we analyse the models' predictions for the invention of novel labels. 

Such errors occur for two out of the four datasets, EURLEX and UKLEX, but with extremely low frequency: the highest observed rate is 0.2\% of novel labels generated for UKLEX (L2). Some examples include `accommodation', `domestic violence' and `vulnerable persons'. Labels in UKLEX and EURLEX are phrased in common terms, compared to the rather technical, domain-specific labels in MIMIC and BIOASQ (see Appendix~\ref{sec:dataset_descr} for examples). Models trained on UKLEX and EURLEX therefore seem to interpret the output space as open-ended and on occasion generate novel labels. Still the total number of novel labels generated is negligible, so this could not explain the lower performance of this approach compared to T5Enc. The reason may instead lie with the fact that Seq2Seq models have to learn the bounds of the output space during training, whereas T5Enc models have that as a given via the fixed decoder input.

\section{Conclusions}
In this work, we compared four approaches to multi-label classification, two based on an encoder only and two based on an encoder-decoder. We experimented with 4 datasets from 2 different domains (legal and biomedical), which support two different label granularities. We found that encoder-decoder methods outperform encoder-only methods, in line with findings in other NLP tasks. We further found that the non-autoregressive use of an encoder-decoder model performs better than using it for conditional generation. We found that decoder depth, width and self-attention are all key contributors to the success of this best approach. 

In future work, we will consider prompt-based approaches as well, specifically instruction-based fine-tuned models~\cite{wei2022finetuned}, currently limited by the excessive computational cost of encoding the full label set as part of the input string.

\section*{Limitations}

Recent work has shown that models of a certain size (upwards of 3B parameters) exhibit learning properties that cannot be observed in smaller models. Due to practical limitations and environmental concerns, in this study we chose not to train models larger than T5-Large. It is thus not possible to know how emergent properties in larger models may have affected the comparison between the different approaches compared here. We believe that our findings will nevertheless be useful to NLP practitioners who operate on a constrained compute budget and may thus opt for moderately-sized models anyway. 

In this work, we compare encoder-only and encoder-decoder models for multi-label classification. Decoder-only (GPT-like) models~\cite{radford2019language} are omitted since at present there are no decoder-only methods for label classification in the literature. While we could have adapted the Seq2Seq approach in our experiments to operate in a decoder-only context, we deem this unsuitable for the datasets we work with, as they contain long documents which will quickly cause problems for standard decoder-only models like GPT-2.

Our experiments consider datasets from the legal and biomedical domains first and foremost because there are publicly available datasets with hierarchical labelling in these domains, unlike others. Moreover, we believe that working in critical application domains is a worthy purpose and covering two such domains with two different datasets in each domain gives us a good view on how the examined methods are expected to work in such domains. 

\section*{Ethics Statement}
The legal and biomedical fields are both highly sensitive and have high impact on human life. In this work, we have ensured that the data we work with is sourced in compliance with the relevant regulations and are fully anonymized where necessary. The application of multi-label classification to this data carries no obvious risk as it can ease the processing and categorization of documents in these domains, without having any direct impact on individuals involved in legal and medical matters.

\section*{Acknowledgments}
This work was fully funded by the Innovation Fund Denmark (IFD, \url{https://innovationsfonden.dk/en}).

\bibliography{anthology,acl2023}
\bibliographystyle{acl_natbib}
\appendix

\begin{table*}[t]
    \centering
    \resizebox{\textwidth}{!}{
    \begin{tabular}{l|cc|cc|cc|cc}
    \toprule
         \multirow{2}{*}{\bf Method} &  \multicolumn{2}{c|}{\bf UKLEX (L1)} & \multicolumn{2}{c|}{\bf EURLEX (L1)} & \multicolumn{2}{c|}{\bf BIOASQ (L1)} & \multicolumn{2}{c}{\bf MIMIC (L1)} \\
         & \microf & \macrof & \microf & \macrof & \microf & \macrof & \microf & \macrof \\
         \midrule
Encoder+Head & 84.3 $\pm$ \small{0.3} & 80.7 $\pm$ \small{0.8} & 82.9 $\pm$ \small{0.2} & 72.5 $\pm$ \small{0.8} & 86.6 $\pm$ \small{0.0} & 77.1 $\pm$ \small{0.2} & 72.4 $\pm$ \small{0.1} & 65.8 $\pm$ \small{0.9} \\
LWAN & 84.5 $\pm$ \small{0.4} & 81.0 $\pm$ \small{1.1} & 83.0 $\pm$ \small{0.2} & 72.2 $\pm$ \small{0.3} & 86.6 $\pm$ \small{0.0} & 77.1 $\pm$ \small{0.3} & 72.5 $\pm$ \small{0.3} & 66.3 $\pm$ \small{1.2} \\
\midrule
Seq2Seq & 84.2 $\pm$ \small{0.0} & 81.6 $\pm$ \small{0.2} & 82.8 $\pm$ \small{0.1} & 74.3 $\pm$ \small{0.5} & 86.5 $\pm$ \small{0.0} & 77.6 $\pm$ \small{0.2} & 73.2 $\pm$ \small{0.1} & \bf 70.3 $\pm$ \small{0.2} \\
Single-Step T5Enc & \bf 85.1 $\pm$ \small{0.2} & 82.4 $\pm$ \small{0.4} & 83.3 $\pm$ \small{0.2} & 73.8 $\pm$ \small{0.8} & 86.7 $\pm$ \small{0.1} & 77.1 $\pm$ \small{0.4} & 73.1 $\pm$ \small{0.1} & 67.4 $\pm$ \small{1.1} \\
T5Enc & 84.8 $\pm$ \small{0.2} & 81.9 $\pm$ \small{0.5} & \bf 83.6 $\pm$ \small{0.1} & \bf 75.0 $\pm$ \small{0.6} & \bf 87.0 $\pm$ \small{0.0} & 78.1 $\pm$ \small{0.3} & \bf 73.6 $\pm$ \small{0.2} & 69.2 $\pm$ \small{1.5} \\
- No attention & 85.0 $\pm$ \small{0.2} & \bf 82.5 $\pm$ \small{0.1} & 83.5 $\pm$ \small{0.1} & 74.8 $\pm$ \small{0.5} & \bf 87.0 $\pm$ \small{0.1} & \bf 78.3 $\pm$ \small{0.2} & 73.6 $\pm$ \small{0.1} & \bf 69.5 $\pm$ \small{0.4} \\
- Full Attention & 84.7 $\pm$ \small{0.3} & 82.1 $\pm$ \small{0.6} & \bf 83.6 $\pm$ \small{0.1} & \bf 75.0 $\pm$ \small{0.3} & \bf 87.0 $\pm$ \small{0.1} & 78.0 $\pm$ \small{0.3} & 73.3 $\pm$ \small{0.1} & 68.7 $\pm$ \small{1.2} \\
\midrule
\multirow{2}{*}{\bf Method} &  \multicolumn{2}{c|}{\bf UKLEX (L2)} & \multicolumn{2}{c|}{\bf EURLEX (L2)} & \multicolumn{2}{c|}{\bf BIOASQ (L2)} & \multicolumn{2}{c}{\bf MIMIC (L2)} \\
         & \microf & \macrof & \microf & \macrof & \microf & \macrof & \microf & \macrof \\
         \midrule
Encoder+Head & 81.9 $\pm$ \small{0.6} & 72.9 $\pm$ \small{1.3} & 76.2 $\pm$ \small{0.2} & 54.0 $\pm$ \small{1.4} & 73.2 $\pm$ \small{0.1} & 60.8 $\pm$ \small{0.8} & 56.7 $\pm$ \small{0.7} & 22.3 $\pm$ \small{1.2} \\
LWAN & 82.0 $\pm$ \small{0.3} & 72.2 $\pm$ \small{0.6} & 76.3 $\pm$ \small{0.3} & 55.5 $\pm$ \small{0.8} & 73.2 $\pm$ \small{0.1} & 60.5 $\pm$ \small{0.8} & 57.2 $\pm$ \small{0.3} & 24.5 $\pm$ \small{0.4} \\
\midrule
Seq2Seq & 81.2 $\pm$ \small{0.3} & 72.7 $\pm$ \small{1.1} & 75.7 $\pm$ \small{0.1} & 57.2 $\pm$ \small{1.1} & 74.1 $\pm$ \small{0.1} & 64.3 $\pm$ \small{0.2} & 57.5 $\pm$ \small{0.3} & 30.7 $\pm$ \small{1.7} \\
Single-Step T5Enc & \bf 82.6 $\pm$ \small{0.1} & \bf 74.4 $\pm$ \small{0.8} & 76.7 $\pm$ \small{0.2} & 55.8 $\pm$ \small{1.4} & 73.5 $\pm$ \small{0.3} & 61.8 $\pm$ \small{1.1} & 58.3 $\pm$ \small{0.5} & 25.8 $\pm$ \small{0.9} \\
T5Enc & 82.4 $\pm$ \small{0.4} & 74.2 $\pm$ \small{1.0} & \bf 77.1 $\pm$ \small{0.1} & \bf 58.8 $\pm$ \small{1.4} & 75.1 $\pm$ \small{0.0} & \bf 66.3 $\pm$ \small{0.1} & \bf 60.6 $\pm$ \small{0.1} & 31.1 $\pm$ \small{1.0} \\
- No attention & 81.9 $\pm$ \small{0.1} & 73.0 $\pm$ \small{0.5} & 76.8 $\pm$ \small{0.1} & 57.6 $\pm$ \small{0.8} & 74.3 $\pm$ \small{0.1} & 64.3 $\pm$ \small{0.3} & 58.6 $\pm$ \small{0.3} & 27.4 $\pm$ \small{1.6} \\
- Full attention & 82.3 $\pm$ \small{0.2} & 74.1 $\pm$ \small{0.8} & \bf 77.1 $\pm$ \small{0.2} & 58.7 $\pm$ \small{0.8} & \bf 75.2 $\pm$ \small{0.0} & 66.1 $\pm$ \small{0.0} & \bf 60.6 $\pm$ \small{0.2} & \bf 31.6 $\pm$ \small{0.7} \\
 \bottomrule
    \end{tabular}
    }
    \caption{Development Results for all methods across datasets with T5 (base).}
    \label{tab:overall_t5_dev}
\end{table*}

\begin{table*}[t]
    \centering
    \resizebox{0.9\textwidth}{!}{
    \begin{tabular}{l|cc|cc|cc|cc}
    \toprule
         \multirow{2}{*}{\bf Method} &  \multicolumn{2}{c|}{\bf UKLEX (L2)} & \multicolumn{2}{c|}{\bf EURLEX (L2)} & \multicolumn{2}{c|}{\bf BIOASQ (L2)} & \multicolumn{2}{c}{\bf MIMIC (L2)} \\
         & \microf & \macrof & \microf & \macrof & \microf & \macrof & \microf & \macrof \\
         \midrule
        \multicolumn{9}{c}{T5 (Small) models} \\
        \midrule
LWAN & 75.1 $\pm$ \small{0.3} & 63.5 $\pm$ \small{0.3} & 69.4 $\pm$ \small{0.2} & 45.0 $\pm$ \small{0.6} & 71.4 $\pm$ \small{0.1} & 56.0 $\pm$ \small{0.2} & 54.4 $\pm$ \small{0.2} & 18.7 $\pm$ \small{0.6} \\
Seq2Seq & 74.0 $\pm$ \small{0.4} & 64.7 $\pm$ \small{0.5} & 68.9 $\pm$ \small{0.6} & 48.7 $\pm$ \small{1.9} & 72.2 $\pm$ \small{0.1} & 60.7 $\pm$ \small{0.2} & 57.8 $\pm$ \small{0.3} & \bf 27.1 $\pm$ \small{0.3} \\
T5Enc & \bf 75.8 $\pm$ \small{0.2} & \bf 65.8 $\pm$ \small{0.4} & \bf 71.4 $\pm$ \small{0.4} & \bf 50.6 $\pm$ \small{1.6} & \bf 73.7 $\pm$ \small{0.1} & \bf 62.4 $\pm$ \small{0.6} & \bf 58.8 $\pm$ \small{0.3} & 25.2 $\pm$ \small{0.4} \\
 \midrule
 \multicolumn{9}{c}{T5 (Large) models} \\
 \midrule
LWAN & 77.1 $\pm$ \small{0.1} & 65.4 $\pm$ \small{0.8} & 70.9 $\pm$ \small{0.1} & 49.4 $\pm$ \small{1.8} & 74.0 $\pm$ \small{0.2} & 61.4 $\pm$ \small{0.9} & 58.3 $\pm$ \small{0.9} & 24.0 $\pm$ \small{3.0} \\
Seq2Seq & 76.5 $\pm$ \small{0.3} & 67.1 $\pm$ \small{0.3} & 71.3 $\pm$ \small{0.1} & 54.1 $\pm$ \small{0.6} & 74.7 $\pm$ \small{0.1} & 65.5 $\pm$ \small{0.5} & 60.4 $\pm$ \small{0.1} & \bf 34.5 $\pm$ \small{0.7} \\
T5Enc & \bf 77.7 $\pm$ \small{0.2} & \bf 68.1 $\pm$ \small{0.7} & \bf 72.4 $\pm$ \small{0.2} & 53.6 $\pm$ \small{1.2} & \bf 75.8 $\pm$ \small{0.1} &\bf 67.1 $\pm$ \small{0.2} & \bf 60.8 $\pm$ \small{0.2} & 33.2 $\pm$ \small{1.6}  \\
 \bottomrule
    \end{tabular}
    }
    \caption{Test Results for all methods across datasets with T5 (small) and (large).}
    \label{tab:overall_t5_small_large}
\end{table*}

\begin{figure*}[!]
\begin{subfigure}{.254\textwidth}
  \centering
  \includegraphics[width=\linewidth]{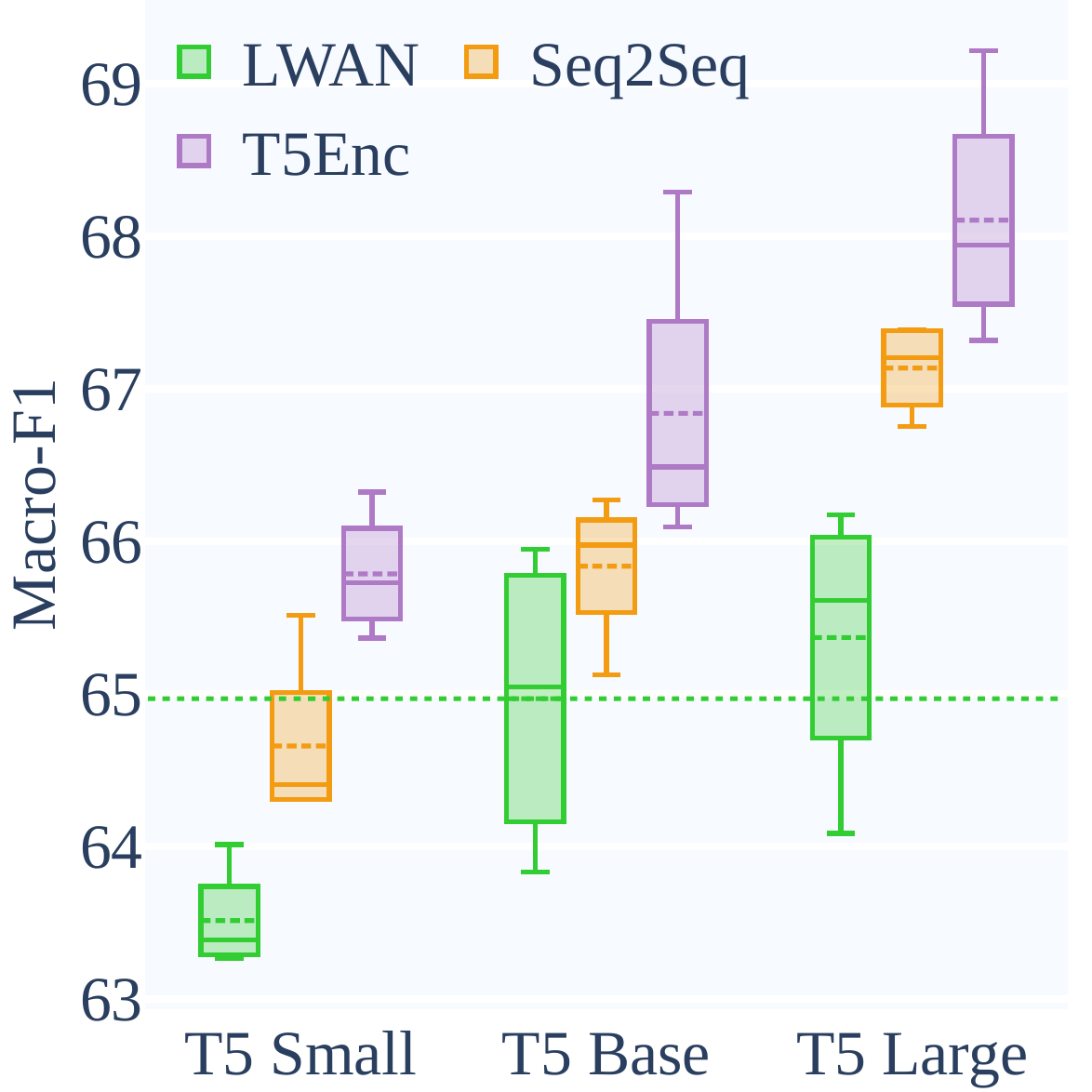}  
  \caption{UKLEX}
  \label{fig:sub-first2}
\end{subfigure}
\begin{subfigure}{.2355\textwidth}
  \centering
  \includegraphics[width=\linewidth]{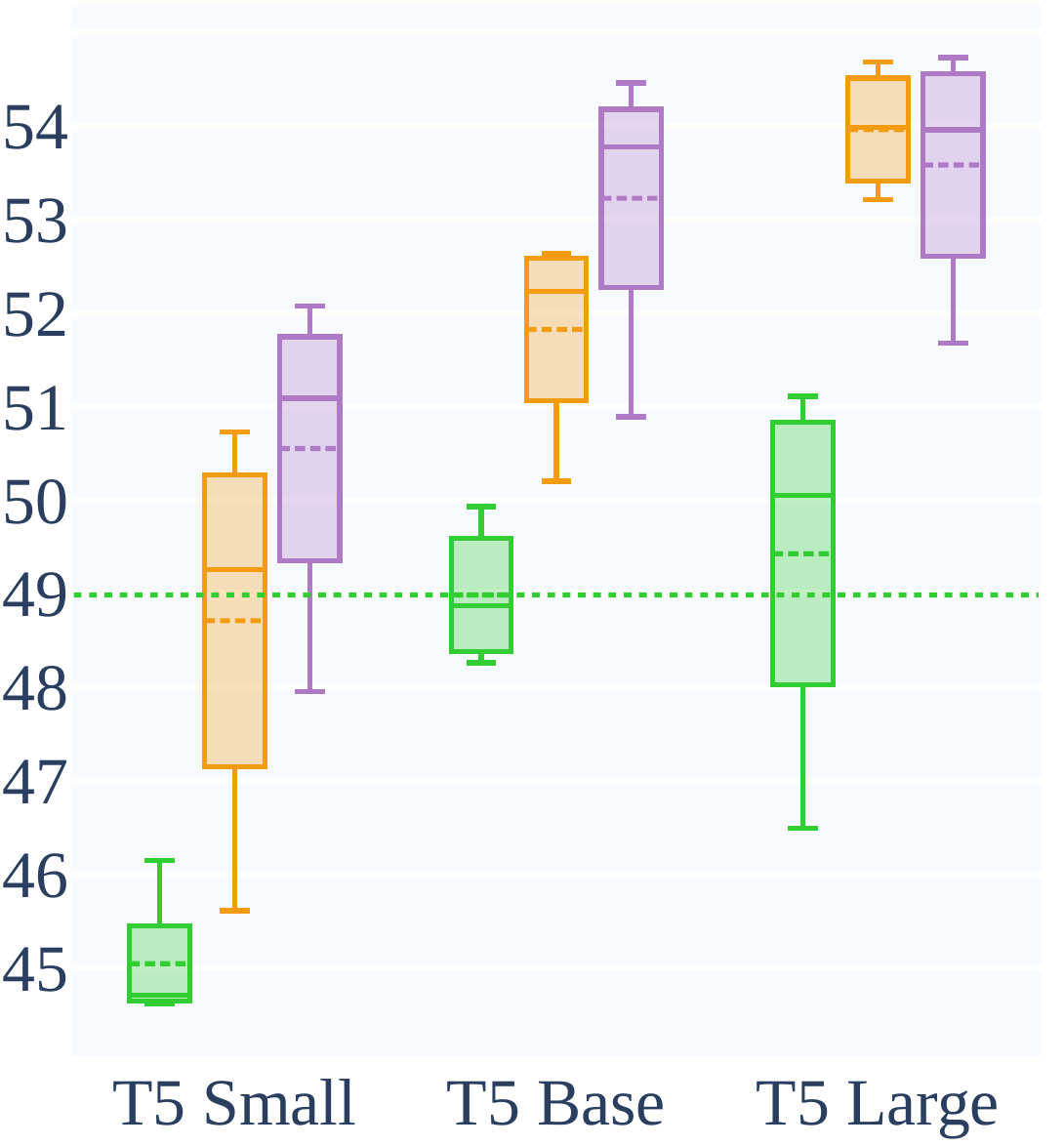}  
  \caption{EURLEX}
  \label{fig:sub-second2}
\end{subfigure}
\begin{subfigure}{.2355\textwidth}
  \centering
  \includegraphics[width=\linewidth]{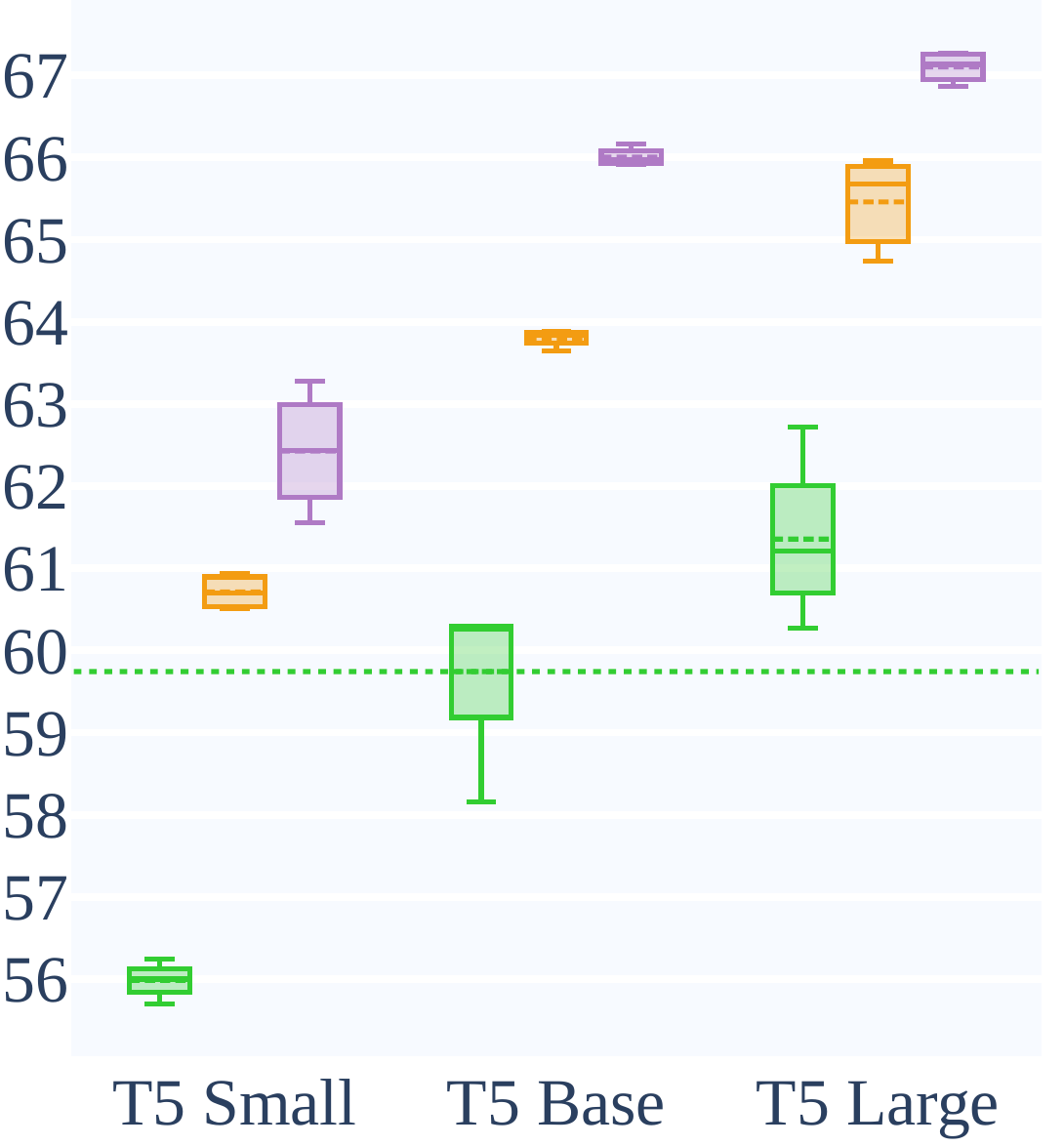}  
  \caption{BIOASQ}
  \label{fig:sub-third2}
\end{subfigure}
\begin{subfigure}{.2355\textwidth}
  \centering
  \includegraphics[width=\linewidth]{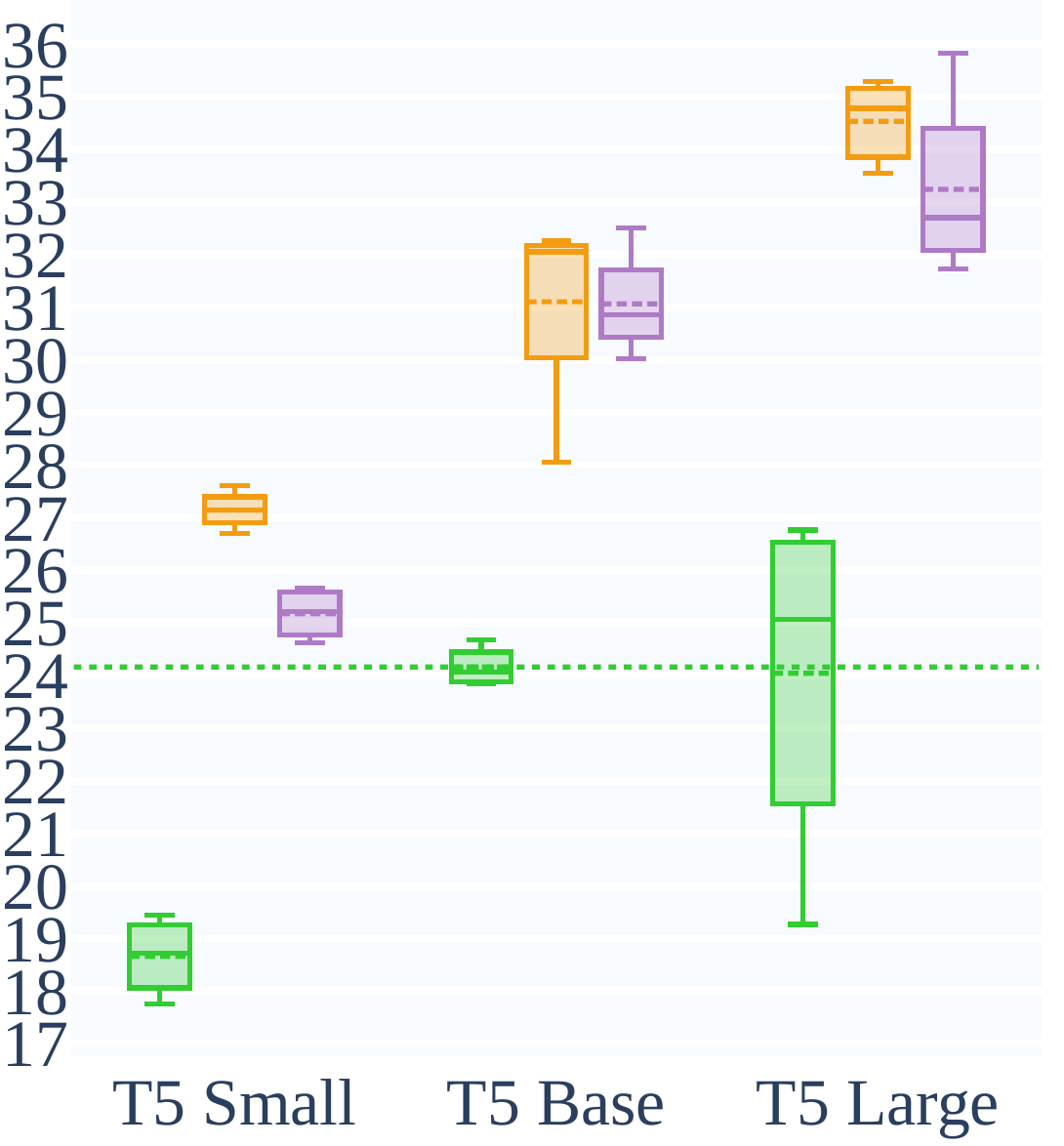}  
  \caption{MIMIC}
  \label{fig:sub-fourth2}
\end{subfigure}
\caption{Performance of the three strongest classification methods across three model sizes in terms of macro-F1 score. Dashed lines within the boxes represent the mean performance across four seeds.}
\label{fig:model_size2}
\end{figure*}

\section{Additional Results}
\label{sec:additional_results}

In Table~\ref{tab:overall_t5_dev}, we present the development results for all models trained from T5-Base across all datasets. 

In Table~\ref{tab:overall_t5_small_large}, we present detailed results for all L2 datasets for methods using T5-Small and Large. In Figure~\ref{fig:model_size2}, which visualizes the macro-F1 results, we see that in comparisons between similarly-sized T5Enc models and LWAN models, the same trends hold here as observed in Section~\ref{sec:experiments} for the micro-F1 metric: T5Enc is superior to LWAN as a method for multi-label classification. 

Comparing Seq2Seq models to similarly-sized LWAN models, on the other hand, we see quite a different trend here compared to the micro-F1 results discussed in Section~\ref{sec:experiments}: Seq2Seq-Small outperforms LWAN-Base models on 2 out of 4 datasets (BIOASQ and MIMIC), and Seq2Seq-Base models outperform LWAN-Base models on all 4 datasets. This suggests that the Seq2Seq approach is especially suitable for the prediction of rare labels, which are better represented by the macro-F1 metric and particularly abundant in the BIOASQ and MIMIC datasets. We presume that as the only approach with access to the actual tokens comprising Level 2 label descriptors, Seq2Seq gains from lexical overlap between label descriptors and prior knowledge of the semantics of these tokens.

In Table~\ref{tab:fisher2}, we show Fisher's exact test results for pair-wise association among labels in Level 1 label sets across all datasets. We see higher rates of pair-wise association, likely because of the smaller number of labels in each set.

\begin{table}[]
    \centering
    \resizebox{\linewidth}{!}{
    \begin{tabular}{l|cccc}
    \toprule
         \bf Level & \textbf{\textbf{\textbf{\textbf{UKLEX}}}} & \textbf{\textbf{\textbf{EURLEX}}} & \textbf{\textbf{BIOASQ}} & \textbf{MIMIC} \textbf{\textbf{}}\\
         \midrule
         L1 & 72.8 & 82.3 & 93.8 & 85.6\\
    
    \bottomrule
    \end{tabular}}
    \caption{Percentage of Level 1 label pairs with significant association according to Fisher's exact test.}
    \label{tab:fisher2}
\end{table}

\section{Dataset descriptors}
\label{sec:dataset_descr}

In Tables~\ref{tab:uklex}, \ref{tab:eurlex}, \ref{tab:bioasq}, \ref{tab:mimic}, we list the original Level 1 and Level 2 label descriptors for the UKLEX, EURLEX, BIOASQ and MIMIC datasets, respectively, as well as the simplified Level 1 label descriptors, which we manually curated.

\begin{table*}[h!]
    \centering
     \resizebox{0.9\textwidth}{!}{
    \begin{tabular}{l|l|l|l|l}
    \toprule
    \toprule
    \multicolumn{5}{c}{\sc Level 1 (Original)} \\
    \midrule
    \midrule
    Agriculture and Food & Children & Criminal Law & Education & Environment \\
    \midrule
    EU & Finance & Healthcare & Housing & Immigration and Citizenship \\
    \midrule
    Local Government & Planning and Development & Politics & Public Order & Social Security \\
    \midrule
    Taxation & Telecommunications & Transportation & - & -\\
    \midrule
    \midrule
    \multicolumn{5}{c}{\sc Level 1 (Simplified)} \\
    \midrule
    \midrule
    Agriculture & Children & Crime & Education & Environment \\
    \midrule
    EU & Finance & Healthcare & Housing & Immigration \\
    \midrule
    Local & Planning & Politics & Public & Social \\
    \midrule
    Taxation & Telecom & Transport & - & -\\
    \midrule
    \midrule
    \multicolumn{5}{c}{\sc Level 2 (Original)} \\
    \midrule
    \midrule
    Agriculture & Air Transport & Animals & Banking & Broadcasting \\ 
    \midrule
    Children & Citizenship & Disabled Persons & Education & Elections \\ 
    \midrule
    Employment & Environment & EU & Finance & Fire and Rescue Services \\ 
    \midrule
    Food & Healthcare & Housing & Immigration & Insurance \\ 
    \midrule
    Land Registration & Local Government & NHS & Police & Pollution \\
    \midrule
    Social Security & Taxation & Telecommunications & Terrorism & Urban Development \\
    \bottomrule
    \end{tabular}
    }
    \caption{Sample of label descriptors (Law Subject)  for UKLEX dataset.}
    \label{tab:uklex}
\end{table*}

\begin{table*}[]
    \centering
    \resizebox{\textwidth}{!}{
    \begin{tabular}{l|l|l|l|l}
    \toprule
    \toprule
    \multicolumn{5}{c}{\sc Level 1 (Original)} \\
    \midrule
    \midrule
    Politics  & European Union & International Relations  & Law  & Economics \\
    \midrule
    Trade  & Finance  & Social Questions  & Education \& Communications  & Science \\
    \midrule
    Business \& Competition & Environment   & Transport  & Working Conditions  & Agriculture  \\ 
    \midrule
    Forestry \& Fisheries  & Agri-Foodstuffs  & Production  & Technology \& Research  & Energy \\
    \midrule
    Industry  & Geography  & International Organisations  & - & -\\
    \midrule
    \midrule
    \multicolumn{5}{c}{\sc Level 1 (Simplified)} \\
    \midrule
    \midrule
    Politics & International & EU & Law & Economy \\ 
    Trade & Finance & Social & Education & Science \\
    Business & Environment & Transport & Employment & Agriculture \\
    Forestry & Food & Production & Technology & Energy \\
    Industry & Geography & Organisations  & - & -\\
    \midrule
    \midrule
    \multicolumn{5}{c}{\sc Level 2 (Original)} \\
    \midrule
    \midrule
    Political Framework & Political Party & Agricultural Activity & Engineering & European Organisations \\
    \midrule
    Politics \& Public Safety & Forestry & International Affairs & Cooperation Policy & International Security \\ 
    \midrule
    Defence & Energy Policy & European Construction & EU Finance & Agricultural Production \\
    \midrule
    Justice & International Law & Rights and Freedoms & Economic Policy & Regional Policy \\
    \midrule
    Economic Structure & Trade Policy & Tariff Policy & International Trade & Marketing \\ 
    \midrule
    Distributive Trades & Monetary Relations & Monetary Economics & Farming Systems & Food Technology \\
    \bottomrule
    \end{tabular}
    }
    \caption{Sample of label descriptors (EUROVOC concepts)  for EURLEX dataset.}
    \label{tab:eurlex}
\end{table*}

\begin{table*}[]
    \centering
     \resizebox{\textwidth}{!}{
    \begin{tabular}{p{5cm}|p{5cm}|p{3cm}|p{5cm}|p{5cm}}
    \toprule
    \toprule
    \multicolumn{5}{c}{\sc Level 1 (Original)} \\
    \midrule
    \midrule
    Anatomy & Organisms & Diseases & Chemicals and Drugs & Analytical, Diagnostic and Therapeutic Techniques and Equipment \\ 
    \midrule
    Psychiatry and Psychology & Phenomena and Processes & Humanities & Disciplines and Occupations &  Anthropology, Education, Sociology, and Social Phenomena\\
    \midrule
    Information Science & Named Groups & Health Care & Technology, Industry, and Agriculture & Publication Characteristics \\
    \midrule
    Geographicals & - & - & - & - \\
    \midrule
    \midrule
    \multicolumn{5}{c}{\sc Level 1 (Simplified)} \\
    \midrule
    \midrule
    Anatomy & Organism & Disease & Drug & Technical \\ 
    \midrule
    Psychology & Process & Occupation & Human & Social \\ 
    \midrule
    Information & Groups & Healthcare & Technology & Publications\\
    \midrule
    Geography & - & - & - & - \\
    \midrule
    \midrule
    \multicolumn{5}{c}{\sc Level 2 (Original)} \\
    \midrule
    \midrule
    Musculoskeletal System & Digestive System & Respiratory System & Urogenital System & Endocrine System \\
    \midrule
    Cardiovascular System & Nervous System & Sense Organs & Embryonic Structures & Cells, Fluids and Secretions \\
    \midrule
    Stomatognathic System & Hemic and Immune Systems & Tissues & Integumentary System & Plant Structures \\ 
    \midrule
    Fungal Structures & Bacterial Structures & Viral Structures & Biomedical and Dental Materials & Microbiological Phenomena \\
    \midrule
   Equipment and Supplies &  Psychological Phenomena & Dentistry &  Mental Disorders Behavior and Behavior Mechanisms &\\
    \bottomrule
    \end{tabular}
    }
    \caption{Sample of label descriptors (MeSH concepts)  for BIOASQ dataset.}
    \label{tab:bioasq}
\end{table*}

\begin{table*}[]
    \centering
    \resizebox{\textwidth}{!}{
    \begin{tabular}{p{5cm}|p{5cm}|p{5cm}|p{5cm}|p{2.5cm}}
    \toprule
    \toprule
    \multicolumn{5}{c}{\sc Level 1 (Original)} \\
    \midrule
    \midrule
    Infection and Parasitic Diseases & Diseases of The Genitourinary System & Endocrine Nutritional and Metabolic Diseases and Immunity Disorders & Diseases of Blood and Blood Forming Organs & Mental Disorders \\
    \midrule
    Diseases of Nervous System and Sense Organs & Diseases of The Circulatory System & Diseases of The Respiratory System & Diseases of The Digestive System &  Neoplasms  \\
    \midrule
    Complications of Pregnancy, Childbirth and the Puerperium & Diseases of The Skin and Subcutaneous Tissue & Diseases of The Musculoskeletal System and Connective Tissue & Certain Conditions Originating In The Perinatal Period & Congenital Anomalies  \\
    \midrule
    Symptoms, Signs and Ill-Defined Conditions & Injury and Poisoning Supplementary Factors Influencing Health Status and Contact With Health Services & Supplementary Classification of External Causes of Injury and Poisoning & - & - \\
    \midrule
    \midrule
    \multicolumn{5}{c}{\sc Level 1 (Simplified)} \\
    \midrule
    \midrule
    Infections & Cancer & Metabolic & Blood & Mental \\
    \midrule
    Nervous & Circular & Respiratory & Digestive & Urinar \\
    \midrule
    Pregnancy & Skin & Muscle & Birth & Newborn \\
    \midrule
    Symptoms & Injury & External & - & - \\
    \midrule
    \midrule
    \multicolumn{5}{c}{\sc Level 2 (Original)} \\
    \midrule
    \midrule
    Osteopathies, Chondropathies, and Acquired Musculoskeletal Deformities  & Bulbus Cordis Anomalies and Anomalies of Cardiac Septal Closure  & Hereditary and Degenerative Diseases of The Central Nervous System  & Poliomyelitis and Other Non-Arthropod-Borne Viral Diseases of Central Nervous System & Tuberculosis \\
    \midrule
    Viral Diseases Accompanied By Exanthem & Arthropod-Borne Viral Diseases & Rickettsioses and Other Arthropod-Borne Diseases & Syphilis and Other Venereal Diseases & Mycoses \\
    \midrule
    Hereditary Hemolytic Anemias & Acquired Hemolytic Anemias & Aplastic Anemia and Other Bone Marrow Failure Syndromes & Other and Unspecified Anemias & Coagulation Defects \\
    \midrule
    Personality Disorders, and Other Nonpsychotic Mental Disorders & Congenital Anomalies of Eye & Inflammatory Diseases of The Central Nervous System & Human Immunodeficiency Virus  & Neurotic Disorders \\
    \midrule
    Disorders of The Peripheral Nervous System & Disorders of The Eye and Adnexa & Diseases of The Ear and Mastoid Process & Chronic Rheumatic Heart Disease & Acute Rheumatic Fever  \\
    \midrule
    Ischemic Heart Disease & Diseases of Pulmonary Circulation & Acute Respiratory Infections & Chronic Obstructive Pulmonary Disease and Allied Conditions & Pneumonia and Influenza  \\
    \midrule
    Intestinal Infectious Diseases & Anencephalus and Similar Anomalies & Other Congenital Anomalies of Nervous System & Zoonotic Bacterial Diseases & Intellectual Disabilities   \\
    \bottomrule
    \end{tabular}
    }
    \caption{Sample of label descriptors (ICD-9 codes) for MIMIC dataset.}
    \label{tab:mimic}
\end{table*}

\end{document}